\documentclass[letterpaper, 10 pt, conference]{ieeeconf}  % Comment this line out if you need a4paper
\usepackage{graphicx} % for pdf, bitmapped graphics files
\usepackage{amssymb}
\usepackage{amsmath}

\IEEEoverridecommandlockouts                              % This command is only needed if 
\title{\LARGE \bf
Sim2Real 3D Object Classification using Spherical Kernel Point Convolution and a Deep Center Voting Scheme}

\author{Jean-Baptiste Weibel$^{1}$, Timothy Patten$^{1,2}$, and Markus Vincze$^{1}$%
\thanks{The research leading to these results has received funding from the Austrian Science Foundation (FWF) under grant agreement No. I3968-N30 HEAP and No. I3969-N30 InDex}
\thanks{$^{1}$Vision for Robotics Laboratory, 
Automation and Control Institute, 
TU Wien, Austria. \{weibel, patten, vincze\}@acin.tuwien.ac.at}

\thanks{$^{2}$Robotics Institute, Faculty of Engineering and Information Technology, University of Technology Sydney, Australia}%

}

\begin{document}

\maketitle
\thispagestyle{empty}
\pagestyle{empty}

\begin{abstract}
While object semantic understanding is essential for most service robotic tasks, 3D object classification is still an open problem. Learning from artificial 3D models alleviates the cost of annotation necessary to approach this problem, but most methods still struggle with the differences existing between artificial and real 3D data. 
We conjecture that the cause of those issue is the fact that many methods learn directly from point coordinates, instead of the shape, as the former is hard to center and to scale under variable occlusions reliably. 
We introduce spherical kernel point convolutions that directly exploit the object surface, represented as a graph, and a voting scheme to limit the impact of poor segmentation on the classification results. Our proposed approach improves upon state-of-the-art methods by up to 36\% when transferring from artificial objects to real objects.

\end{abstract}

%%%%%%%%%%%%%%%%%%%%%%%%%%%%%%%%%%%%%%%%%%%%%%%%%%%%%%%%%%%%%%%%%%%%%%%%%%%%%%%%
%%%%%%%%%%%%%%%%%%%%%%%%%%%%%%%%%%%%%%%%%%%%%%%%%%%%%%%%%%%%%%%%%%%%%%%%%%%%%%%%
\section{Introduction}

%%% Vision:
Service robotics could provide significant relief to overwhelmed healthcare workers or help elderly people remain independent for longer by carrying out menial tasks. To attain this objective, they need to adapt to an ever changing environment and manipulate a constantly evolving set of objects. The majority of state-of-the-art methods use Deep Learning to achieve this property~\cite{dgcnnwang2019, pointcnnli2018, spidercnnxu2018, pointnet++qi2017}. However, such approaches require a large constant flow of annotated data to learn about new objects, which is only practical through simulation and modeling, as manual reconstruction, segmentation and labeling at scale is too time consuming.

%%% Challenge:
Learning a model applicable on real data from artificially produced data is challenging. The loss of performance, usually referred to as the Sim2Real gap, is paradoxically larger in 3D object classification than it is for 2D color images~\cite{scanobjectnn-iccv19}, even though producing photo-realistic images is more challenging than producing realistic 3D models. 
Notably, scale from CAD models is arbitrary, and not necessarily metric, making it difficult to rely on when using a large-scale database. In combination with commonly observed occlusions, it becomes challenging to obtain consistent coordinates for object points, as centering and scaling operations will be affected by the occlusions themselves. In addition to occlusions, objects are typically poorly segmented. Relying explicitly or implicitly on coordinates render those representation more susceptible to misclassification on real data.

\begin{figure}[t]
   \centering
   \includegraphics[scale=0.65]{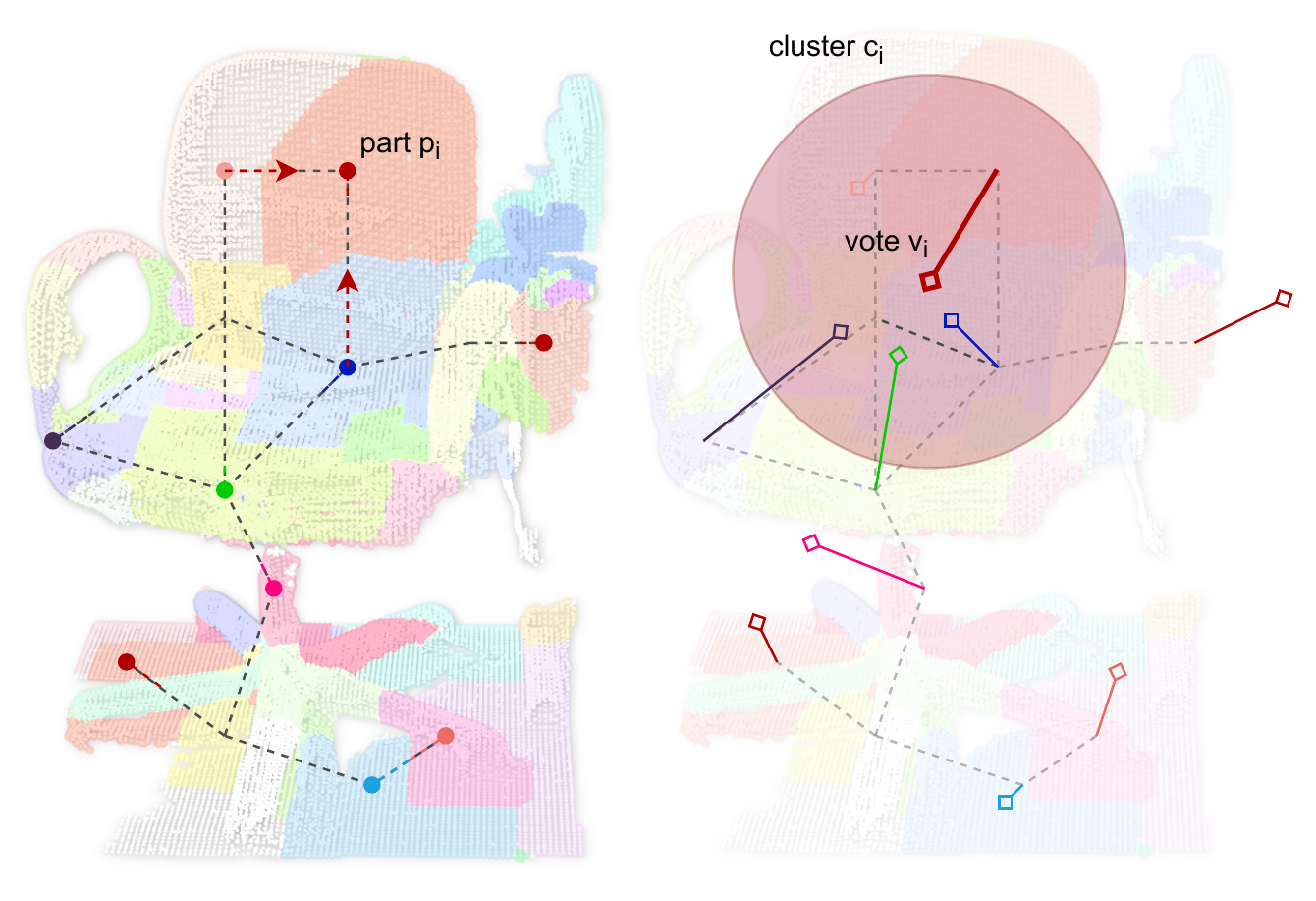}
   \caption{Illustration of the voting procedure. Every part of the graph learns from its own shape and its neighborhood (left). They each cast a vote for the object center that then get clustered (right). Votes from background points are more scattered than votes from the object, making the separation easier.}
   \label{fig:teaser}
\end{figure}

%%% Approach:
We propose a novel architecture that creates scale-invariant and rotation-invariant representations called Spherical Kernel Point Convolution (SKPConv) that does not rely on a global coordinate frame. Representations are learned from a graph of parts, only considering the shape of parts, their connectivity, and relative orientation instead of point coordinates. In addition, we introduce a voting mechanism to further robustify our method to under-segmentation as illustrated in Figure~\ref{fig:teaser}. Each part votes for its corresponding object center, and background and foreground parts are separated through their center prediction.

More specifically, our contributions are:

\begin{itemize}
    \item A novel graph convolution layer called SKPConv focusing on the relative orientation of nodes based on the kernel point convolution.
    \item The introduction of a voting procedure for foreground / background separation.
\end{itemize}

%%% Results preview:
When learning on ModelNet artificial models, our approach improves upon state-of-the-art on the ScanObjectNN~\cite{scanobjectnn-iccv19} real object dataset by up to 36\% on its noisiest variation.

%%% Outline:
The remainder of the paper is organized as follows. Section~\ref{sec:related} reviews the recent related literature. Section~\ref{sec:method} introduces our novel approach dealing with the Sim2Real gap. Finally, section~\ref{sec:exps} presents our results on noisy occluded and under-segmented point cloud from the ScanObjectNN dataset~\cite{scanobjectnn-iccv19}.

%%%%%%%%%%%%%%%%%%%%%%%%%%%%%%%%%%%%%%%%%%%%%%%%%%%%%%%%%%%%%%%%%%%%%%%%%%%%%%%%
%%%%%%%%%%%%%%%%%%%%%%%%%%%%%%%%%%%%%%%%%%%%%%%%%%%%%%%%%%%%%%%%%%%%%%%%%%%%%%%%
\section{Related work}\label{sec:related}
This section introduces methods for 3D data object classification, first focusing on methods with a strong dependence on point coordinates, then on methods taking advantage of a graph structure over points and finally methods specifically designed for robustness against difficulties commonly found in real data.

\subsection{Grid-based classification of 3D data}

While a seemingly natural extension of the Deep Learning methods created for images, voxel-grid based methods~\cite{voxnetmaturana2015} struggle to compete with more novel approaches, as the increase in memory required by the additional dimension generally leads to the use of coarser and less representative grids. 
One of the very first deep learning methods designed specifically for point clouds is  PointNet~\cite{qi_pointnet:_2017}. It is invariant to point ordering as it considers one point at a time. The global optimization and the sole reliance on point coordinates before the max-pooling step drives the creation of an implicit grid encoded in the network weights. PointNet++~\cite{pointnet++qi2017} was then introduced as an extension and combines multiple layers of PointNet to increase the representational power of the network. Other methods, such as 3dmFV~\cite{3dmFV-8394990}, create a grid over the unit sphere and accumulate the influence of each point on each area of the grid, without discontinuities that would be introduced by a voxel grid.

\subsection{Neighborhood-based classification of 3D data}

While not naturally present in point clouds, many methods have been developed to learn features based on the neighborhood of each point. The neighborhood information can efficiently be recovered in a point cloud using a KD-Tree. SpiderCNN~\cite{spidercnnxu2018} adapts the regular convolution for point clouds by using a third-order Taylor decomposition to learn surface information. Similarly, PointCNN~\cite{pointcnnli2018} tries to generalize the regular convolution for point clouds. Finally, KPConv~\cite{KPConvthomas2019} instantiates a set of kernels at fixed areas around every point of interest and weights the features of every neighbor to those kernels based on the distance to it. 

\subsection{Graph-based classification 3D data}

Another direction for 3D data classification is to represent the surface of objects as a graph, for example, as 3D meshes, which are graphs with a specific topology. This is a commonplace representation in Computer Graphics. Learning from such a representation focuses on the continuous surface and the structure instead of relying on point coordinates. 
The graph convolutional network~\cite{kipf2017semi} and its more powerful variant~\cite{gatvelickovic2018graph} are two general architectures designed to learn from graphs. These works have been adapted for 3D classification in ~\cite{verma_feastnet:_2018} and~\cite{weibel2019addressing}. Other approaches learn from the edges of the graph, while dynamically updating the graph~\cite{dgcnnwang2019}. Only considering surfaces as a general graph loses important information about the relative position of nodes. Mesh-specific architectures also exist~\cite{Gong_2019_ICCV} but tend to create an over-reliance on the structure itself, becoming sensitive to slight variation in position and triangulation.

\subsection{Robust representations for 3D data}
Finally, a large set of methods has been created or adapted to be robust to specific forms of noise. \cite{scanobjectnn-iccv19} introduces an extra segmentation step to limit the effect of background points in poorly segmented objects but it cannot be trained purely from CAD models. More classical approaches such as the Ensemble of Shape Features, have also proven their resilience to most sources of noise found in real 3D data~\cite{wohlkinger_ensemble_2011}~\cite{bobkov2018noise}. Finally, some methods have been designed to be invariant to one specific aspect, such as rotation-invariance using KPConv~\cite{rotkpconv9320370}.

%%%%%%%%%%%%%%%%%%%%%%%%%%%%%%%%%%%%%%%%%%%%%%%%%%%%%%%%%%%%%%%%%%%%%%%%%%%%%%%%
%%%%%%%%%%%%%%%%%%%%%%%%%%%%%%%%%%%%%%%%%%%%%%%%%%%%%%%%%%%%%%%%%%%%%%%%%%%%%%%%
\section{Sim2Real Classification of Under-segmented Objects}\label{sec:method}

The key idea of the proposed method is to avoid the use of any implicit global reference frame, as it would make the resulting representation sensitive to occlusions. To achieve this, we represent the object as a graph of object parts. Its creation and the learning architecture used to preserve those properties is described in the following sections.

\subsection{Building a graph of parts}

Building on the approach of~\cite{weibel2019addressing}, a graph of object parts is created from the raw object mesh or point cloud. In order to obtain similar results independently of the orientation, scale, or occlusion applied to the input points, the parts are grown based on the surface curvature. More specifically, from a randomly sampled point, the part is grown iteratively based on the angle between normals of neighboring points. Points are accumulated until the sum of angles reaches a threshold, or no more points are available. Indeed, only neighboring points are considered when growing the part. A neighborhood is defined by the surface of the mesh, or computed using a KD-Tree in the case of a point cloud, excluding points whose normal is in the opposite direction. A given point can also only be sampled by a single part. To guarantee that nodes in the final graph are well connected, the next part seed is sampled from the borders of the already sampled parts.

\begin{figure}[t]
   \centering
   \includegraphics[scale=0.8]{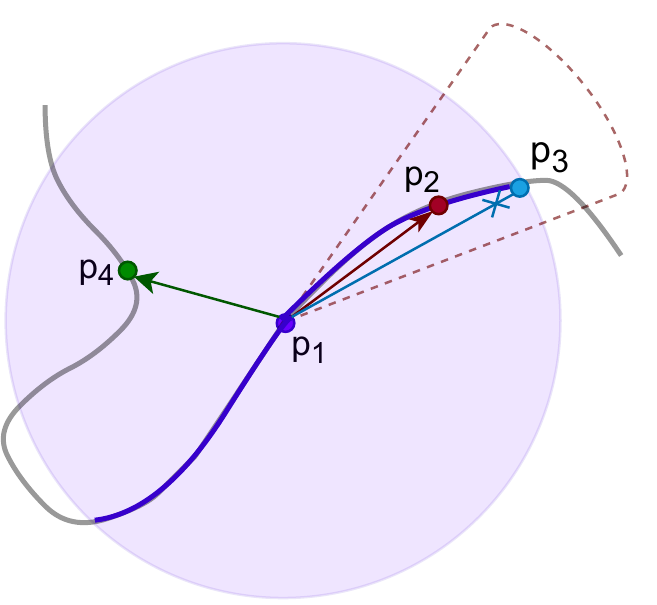}
   \caption{Illustration of the connectivity in the graph in a 2D scenario. The part $p_1$ bounding sphere and elements are represented. It is connected to the part $p_4$ through the KD-Tree, but not part $p_3$, as it is already ``hidden" by the part $p_2$ (materialised by a cone).}
   \label{fig:connectivity}
\end{figure}

The connectivity between the parts is defined by the object surface. However, real scenes tend to have many disconnected large areas. To compensate for this, each part is connected to any other part that intersects with the bounding sphere of the part of interest. Only the closest neighbor in a given direction is kept, as those ``hidden" by a closer neighbor are disconnected. Neighbors in the opposite direction of the part normal are also ignored. This process is illustrated in Figure~\ref{fig:connectivity}

A repeatable local reference frame is computed for each part such that the learned representation can be rotation-invariant. The principal component axis is computed over the set of points of each part, obtaining the eigenvectors $\vec{e_{i,1}}$, $\vec{e_{i,2}}$, and $\vec{e_{i,3}}$ where only $\vec{e_{i,3}}$ (the axis normal to the surface) is kept. From this, the local reference frame is built using the global horizontal axis, and the component of the part normal that is orthogonal to the global horizontal axis. The last vector of the local reference frame is obtained by using the cross-product over the two vectors defined. So, for $\vec{x}$, $\vec{y}$ and $\vec{z}$ the global coordinate frame, our coordinate frame is defined as:
\begin{align*}
    \vec{v_{i, 1}} & = \vec{e_{i, 3}} - \left(\vec{e_{i, 3}} . \vec{z} \right) \vec{z} \\
    \vec{v_{i, 2}} & = \vec{v_{i, 3}} \times \vec{v_{i, 1}} \\
    \vec{v_{i, 3}} & = \vec{z}
\end{align*}

Growing the part based on the angular variation guarantees that no part will be flat, which is beneficial as it prevents the part normal to be co-linear to the global horizontal axis. The only situation where this could happen is if the surface is purely horizontal and captured under a large amount of centered noise. This is unlikely to occur in 3D reconstructions since surfaces (e.g., table, desks or floor) are usually visible from many viewpoints, which reduces the final noise level. In this case, the identity matrix is used as the orientation of the local reference frame. The translation is defined by the part's mean. This guarantees that the representation is always invariant to any rotation around the global horizontal axis.

%%%%%%%%%%%%%%%%%%%%%%%%%%%%%%%%%%%%%%%%%%%%%%%%%%%%%%%%%%%%%%%%%%%%%%%%%%%%%%%%
\subsection{Learning object representations from object graphs}
\subsubsection{Local parts representation}
Parts created using the method described in the previous section can represent very different surfaces and have very different volumes. A part-specific representation is therefore introduced. A fixed number of points within the part is sampled, each point is centered and rotated using the local reference frame described in the previous section and the set of points of the part is scaled to fit the unit sphere. The parts representation is learned using an architecture similar to PointNet~\cite{qi_pointnet:_2017}, except that at every one-dimensional convolution layer, the maximum of the input feature is subtracted, weighted by a learned parameter as described in~\cite{iclr_2017_dl_sets_pc}, and batch normalization~\cite{batchnorm_ioffe_2015} is used at every layer.

%%%%%%%%%%%%%%%%%%%%%%%%%%%%%%%%%%%%%%%%%%%%%%%%%%%%%%%%%%%%%%%%%%%%%%%%%%%%%%%%
\subsubsection{Representing parts in context using Spherical Kernel Point Convolutions}

Learning directly from the structure of the graph instead of relying on point coordinates at a larger scale avoids the use of any coordinate frame, which cannot be reliably obtained from variably occluded objects under different scales. However, graph convolutions as introduced in~\cite{kipf2017semi} do not learn anything from the relative orientation between parts and focus on their connectivity. Kernel point convolutions~\cite{KPConvthomas2019} can learn from both aspects. For point coordinates $p_i \in \mathbb{R}^3$ and their corresponding features $f_i \in \mathbb{R}^{F_{in}}$ where $F_{in}$ is the dimension of the input features, the convolution of kernel $g$ is defined as:

\begin{equation}
    f_i * g = \sum_{x_j \in \mathcal{N}_i} g(p_j, p_i)  f_j
\end{equation}

To introduce different weights depending on the relative position of points, as in convolutions over grids, $K$ kernels spread on a sphere around the point of interest are created. The kernels' weights $W_k$ are applied to every neighboring feature, weighted by the influence function of the kernel $h_k$:

\begin{equation}
    g(p_j, p_i) = \sum_{k \in [1,K]} h_k(p_j - p_i) W_k
\end{equation}
This influence function weighs points closer to the kernel center $c_k$ more than those further away:

\begin{equation}
    h_k(p) = \max \left(0, \frac{\left\| p -  c_k \right\|}{\sigma} \right)
\end{equation}

\begin{figure}[t]
   \centering
   \includegraphics[scale=1.]{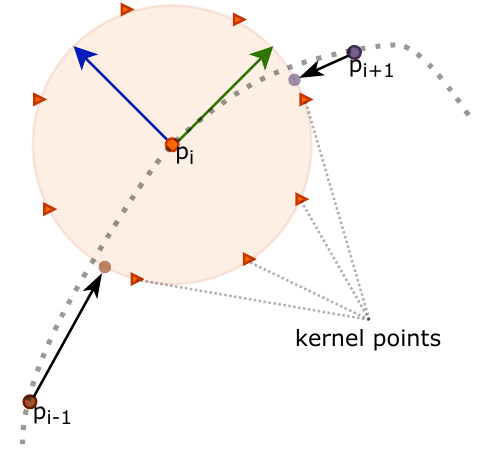}
   \caption{Spherical Kernel Point Convolutions for the point $p_i$ only take into account the direction of neighbors $p_{i-1}$ and $p_{i+1}$  as illustrated in this 2D scenario.}
   \label{fig:skpconv}
\end{figure}

The center of the parts defined in the previous section is used as the input point cloud to the kernel point convolution. This point cloud is sparse and has variable density due to the variable size of parts. For a given point, neighbors might therefore not be close enough to any kernel center and have very little influence on the resulting features. To compensate, we introduce the spherical kernel point convolution, which projects each neighbor on the unit sphere around the point of interest, thus only considering the direction and not the distance of each neighbors. This process is illustrated in Figure~\ref{fig:skpconv}.

To preserve the rotation-invariance property, the neighboring points of each part are transformed using the local reference frame defined in the previous section.

\begin{figure*}[t]
   \centering
   \includegraphics[scale=0.8]{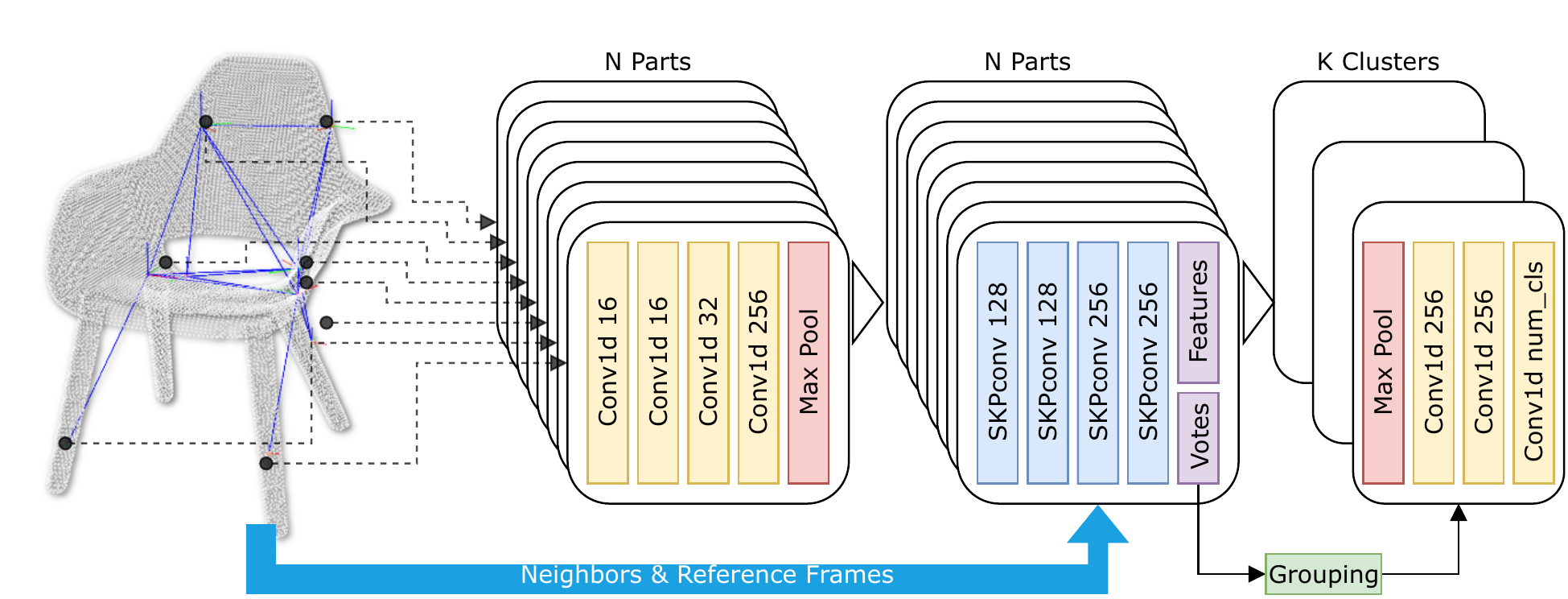}
   \caption{Overview of our approach. A local representation is learned for the N parts extracted from the object. Taking advantage of their connectivity and relative position, four SKPConv layers are used to learn the final part representation and vote for the object center. Parts features are then grouped based on the position of votes to create the final prediction.}
   \label{fig:system}
\end{figure*}

%%%%%%%%%%%%%%%%%%%%%%%%%%%%%%%%%%%%%%%%%%%%%%%%%%%%%%%%%%%%%%%%%%%%%%%%%%%%%%%%
\subsection{Separating objects from the background using a voting scheme}

The architecture introduced so far produces a rotation- and scale-invariant representation, and is robust to occlusion. However, in the context of Sim2Real classification, objects are also often under-segmented, meaning that not every point belongs to the object of interest. As good classification results require good segmentation and good segmentation masks require good classification, we propose to frame this problem of under-segmented object classification as a detection problem. Based on the findings of~\cite{votenetqi2019deep}, we conjecture that background parts and object parts can be separated using object center prediction as object parts will have more consistent votes.

Each part predicts an offset $\Delta p_i$ from the part center in the part local reference frame. All votes are then transformed back to the global reference frame using the inverse rotation defined in the local reference frame $R^{-1}_i$. Using farthest point sampling, a fixed number of cluster centers is sampled among the votes cast. Every vote within a sphere of fixed radius is considered a part of the cluster. Using a max-pooling layer, we aggregate the features of all parts whose vote was cast within a cluster. This cluster representation is then classified. The full architecture is illustrated in Figure~\ref{fig:system}. 

During training, only CAD models are used, which can easily be centered, avoiding the need for any extra annotation. The voting procedure is supervised by an L2-loss:
\begin{equation}
    vote\_loss = \left\| R^{-1}_{i}\Delta p_i + p_i \right\|^2    
\end{equation}

The part representation does not have access to its own coordinates in the global reference frame, which ensures that the voting procedure cannot fall back on a trivial solution. Keeping the whole architecture scale-invariant makes the voting procedure more uncertain, but this remains an effective foreground-background strategy if the object classes considered have sufficiently similar scales.

%%%%%%%%%%%%%%%%%%%%%%%%%%%%%%%%%%%%%%%%%%%%%%%%%%%%%%%%%%%%%%%%%%%%%%%%%%%%%%%%
%%%%%%%%%%%%%%%%%%%%%%%%%%%%%%%%%%%%%%%%%%%%%%%%%%%%%%%%%%%%%%%%%%%%%%%%%%%%%%%%
\section{Experiments}\label{sec:exps}

This section introduces our experimental results. We first consider the performance of our approach on the standard ModelNet benchmark, a set of artificial CAD models. We then present our results when learning from artificial data and testing on real data. Finally, the impact of our design choices are evaluated.

\subsection{Experimental setup}

Experiments are performed using the ModelNet dataset~\cite{modelnet}, a set of 12311 CAD models from 40 classes, and the ScanObjectNN dataset~\cite{scanobjectnn-iccv19}, a set of 2902 unique real object instances. To obtain meshes closer to the ones obtained from real reconstructions from artificial CAD models, a set of views is sampled and fed to a TSDF (Truncated Signed Distance Function) volume. The meshes obtained from those reconstructions are used for the experiments.

The graph of object parts is created with up to 128 parts, and 128 points are sampled within each part. Votes are grouped into 5 clusters. Since the task at hand is classification, only one prediction per object is necessary, so only the predictions from the most confident cluster, as defined by the softmax are used. The number of layers and number of features used at each layer is described in Figure~\ref{fig:system}.

During training, artificial data is augmented in two ways. Random noise is added to the normals to make the size of parts less consistent and closer to parts in real data. Random occlusions are also generated by sampling one normal on the object and only keeping points whose normals are at an angle of less than 90° with the originally sampled point. This simulates varying amount of occlusion based on the object geometry and also creates more realistic occlusions. Indeed, points that could have been visible together in any view used for the object reconstruction will be kept, as their surfaces have similar orientation.

\subsection{Evaluation on artificial data}

In this section, for completeness, the performance of our method is compared to the same selection of methods as in~\cite{scanobjectnn-iccv19} on artificial data, although our goal is to improve the performance on real-world scenes. They are all state-of-the-art methods developed specifically for point clouds. The results on the 40 classes of the ModelNet dataset~\cite{modelnet} are presented in Table~\ref{table:expMN}.  Minor adaptation to our method are done when testing on artificial data: no KD-Tree is used to connect parts, since the surface is reliable and un-occluded in this case. The results presented are otherwise obtained using the same configuration as for the transfer experiment.

Our method is slightly less competitive on artificial data. This is in part due to the way the graph is designed. Parts center are sampled randomly, and without any form of noise on the surface, the final part becomes sensitive to the center. Real data always has some level of noise, paradoxically making this sampling slightly more predictable. 
%Further tuning could improve results but the focus of this work is on the transfer experiment.

\begin{table}[t]
\caption{Classification experiments on ModelNet40~\cite{modelnet}}
\label{table:expMN}
\begin{center}
\begin{tabular}{|c| c| c| }
  \hline
  \textbf{Method} & \textbf{Acc.} & \textbf{Cls Acc.} \\
  \hline
  3DmFV~\cite{3dmFV-8394990} & 91.4 & 86.3 \\
  \hline
  PointNet~\cite{qi_pointnet:_2017} & 89.2 & 86.2 \\
  \hline
  SpiderCNN~\cite{spidercnnxu2018} & 90.0 & 86.8 \\
  \hline
  PointNet++~\cite{pointnet++qi2017} & 90.7 & 87.8 \\
  \hline
  DGCNN~\cite{dgcnnwang2019} & 92.2 & \textbf{90.2} \\
  \hline
  PointCNN~\cite{pointcnnli2018} & \textbf{92.5} & 88.8\\
  \hline
  GAT-PointNet~\cite{weibel2019addressing} & 89.4 & 87.2 \\
  \hline
  Ours & 88.6 & 85.8 \\
  \hline
\end{tabular}
\end{center}
\end{table}

\subsection{Evaluation of the gap between artificial and real data}

\begin{figure}[t]
   \centering
   \includegraphics[scale=0.2]{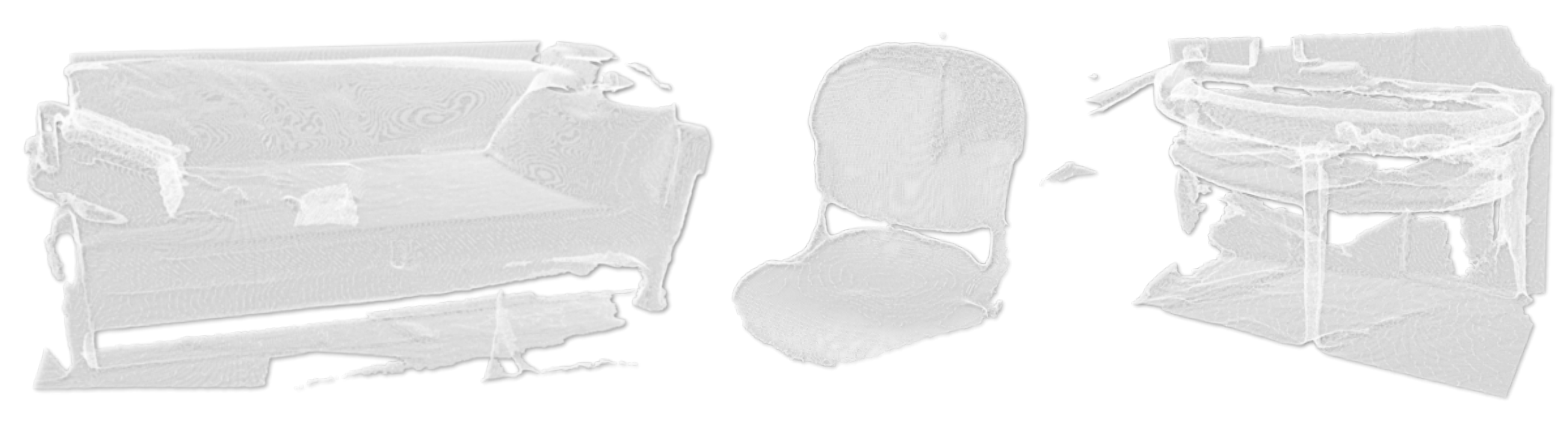}
   \caption{Example objects from ScanObjectNN. Objects have variable level of occlusion and background points, but at least 50\% of the points are part of the object.}
   \label{fig:sonn}
\end{figure}

In this section, models are trained on a subset of ModelNet based on the classes overlapping with the classes defined in ScanObjectNN. The evaluation is then performed on the original version of ScanObjectNN, with and without background points. Background points are points within the bounding box around the original object. The evaluation is also performed on perturbed versions of ScanObjectNN, specifically ScanObjectNN\_T25, where bounding boxes from the original ScanObjectNN are translated by up to 25\% of the object size, and ScanObjectNN\_T50\_RS where bounding boxes are translated by up to 50\% and objects are rescaled and rotated randomly. All results reported are obtained using ScanObjectNN's main train/test split.  More details on the perturbation and split definition can be found in~\cite{scanobjectnn-iccv19}. Example objects are shown in Figure~\ref{fig:sonn}.
The threshold on accumulation for part sampling is set three times higher on real data than on artificial data to compensate for the extra noise.

Table~\ref{table:expMNtoSONNnoBG} and~\ref{table:expMNtoSONNwBG} clearly show the benefit of our approach on the task of transferring from artificial data to real data as it equals or outperforms every other method on every variant of the ScanObjectNN dataset. The more perturbed the input, the larger the improvement compares to other methods. In Table~\ref{table:expMNtoSONNnoBG}, the benefits of our SKPConv layer is quite clear. The increase in accuracy is, respectively, 6.9\%, 11.3\% and 13.3\% over the state-of-the-art for the original, the T25 and T50\_RS variants of ScanObjectNN.

Table~\ref{table:expMNtoSONNwBG} illustrates the combined benefits of our SKPConv layer and our voting scheme as we test on ScanObjectNN while keeping the background points. The difference in accuracy with respect to the state-of-the-art is -0.9\%, 25.2\% and 36.0\%, for the original, T25 and T50\_RS variants of ScanObjectNN. For the original version of ScanObjectNN, our method is only as good as DGCNN~\cite{dgcnnwang2019} when including the background points. DGCNN degrades much better in this specific configuration (5\% decrease in accuracy when including background points) as in any other configuration (21\% decrease on T25 variant and 26\% decrease on T50\_RS when including background points).
    
%\begin{table}[t]
%\caption{Evaluation when training on ModelNet and testing on ScanObjectNN}
%\label{table:expMNtoSONN}

%\begin{center}
%\begin{tabular}{|c| c| c| c| c| c| c| }
%  \hline
  %& \multicolumn{2}{|c|}{\textbf{OBJ}} & \multicolumn{2}{|c|}{\textbf{T25}} & \multicolumn{2}{|c|}{\textbf{T50\_RS}} \\
  %\hline
  %With backg. & N & Y & N & Y & N & Y \\
  %\hline
  %3DmFV~\cite{3dmFV-8394990} & 30.9 & 24.0 & 28.4 & 19.9 & 24.9 & 16.4 \\
  %\hline
  %PointNet~\cite{qi_pointnet:_2017} & 42.3 & 41.1 & 37.6 & 30.1 & 31.1 & 23.2 \\
  %\hline
  %SpiderCNN~\cite{spidercnnxu2018} & 44.2 & 42.1 & 37.7 & 26.8 & 30.9 & 22.2 \\
  %\hline
  %PointNet++~\cite{pointnet++qi2017} & 43.6 & 37.7 & 37.8 & 28.2 & 32 & 22.9 \\
  %\hline
  %DGCNN~\cite{dgcnnwang2019} & 49.3 & \textbf{46.7} & 42.4 & 33.3 & 36.8 & 27.2 \\
  %\hline
  %PointCNN~\cite{pointcnnli2018} & 32.2 & 29.5 & 28.7 & 21.9 & 24.6 & 19.2 \\
  %\hline
  %GAT-PointNet similar to~\cite{weibel2019addressing} & 38.4 & 31.8 & 36.7 & 28.5 & 33.8 & 26.0 \\
  %\hline
  %Ours & \textbf{52.7} & 46.3 & \textbf{47.2} & \textbf{41.7} & \textbf{41.7} & \textbf{37.0} \\
  %\hline
%\end{tabular}
%\end{center}
%\end{table}

\begin{table}[t]
\caption{Evaluation when training on ModelNet and testing on ScanObjectNN without any background points}
\label{table:expMNtoSONNnoBG}
\begin{center}
\begin{tabular}{|c| c| c| c| }
  \hline
  & \textbf{OBJ} & \textbf{T25} & \textbf{T50\_RS} \\
  \hline
  3DmFV~\cite{3dmFV-8394990} & 30.9 & 28.4 & 24.9 \\
  \hline
  PointNet~\cite{qi_pointnet:_2017} & 42.3 & 37.6 & 31.1 \\
  \hline
  SpiderCNN~\cite{spidercnnxu2018} & 44.2 & 37.7 & 30.9 \\
  \hline
  PointNet++~\cite{pointnet++qi2017} & 43.6 & 37.8 & 32.0 \\
  \hline
  DGCNN~\cite{dgcnnwang2019} & 49.3 & 42.4 & 36.8 \\
  \hline
  PointCNN~\cite{pointcnnli2018} & 32.2 & 28.7 & 24.6 \\
  \hline
  GAT-PointNet similar to~\cite{weibel2019addressing} & 38.4 & 36.7 & 33.8 \\
  \hline
  Ours & \textbf{52.7} & \textbf{47.2} & \textbf{41.7} \\
  \hline
\end{tabular}
\end{center}
\end{table}

\begin{table}[t]
\caption{Evaluation when training on ModelNet and testing on ScanObjectNN including background points}
\label{table:expMNtoSONNwBG}
\begin{center}
\begin{tabular}{|c| c| c| c| }
  \hline
  & \textbf{OBJ} & \textbf{T25} & \textbf{T50\_RS} \\
  \hline
  3DmFV~\cite{3dmFV-8394990}  & 24.0 & 19.9 & 16.4 \\
  \hline
  PointNet~\cite{qi_pointnet:_2017} & 41.1 & 30.1 & 23.2 \\
  \hline
  SpiderCNN~\cite{spidercnnxu2018} & 42.1 & 26.8 & 22.2 \\
  \hline
  PointNet++~\cite{pointnet++qi2017} & 37.7  & 28.2 & 22.9 \\
  \hline
  DGCNN~\cite{dgcnnwang2019} & \textbf{46.7} & 33.3 & 27.2 \\
  \hline
  PointCNN~\cite{pointcnnli2018} & 29.5 & 21.9 & 19.2 \\
  \hline
  GAT-PointNet similar to~\cite{weibel2019addressing} & 31.8 & 28.5 & 26.0 \\
  \hline
  Ours & 46.3 & \textbf{41.7} & \textbf{37.0} \\
  \hline
\end{tabular}
\end{center}
\end{table}

\subsection{Ablation study}

In this section, different design choices are evaluated independently and results are presented in Table~\ref{table:expAblation}. Specifically, spherical kernel points convolution (SKPConv) are compared to kernel point convolutions (KPConv)~\cite{KPConvthomas2019}, given the same graph, and to Graph Attention Layers~\cite{gatvelickovic2018graph}, making the overall approach similar to~\cite{weibel2019addressing}. The voting scheme presented in Section~\ref{sec:method} is referred to as VoteMaxPool and is evaluated against max-pooling over the features of every part (MaxPool).

We can see that SKPConv achieves a 66\% increase in accuracy over the regular KPConv when applied to the sparse graph with variable density created by our method. While robust, using GAT is not nearly as accurate, due to the fact that it cannot take advantage of the relative position of nodes in the graph.
The improvement of the VoteMaxPool pooling strategy over the MaxPool shows on objects with background points, with no negative impact on the performance for accurately segmented out objects.

\begin{table}[t]
\caption{Ablation study. Results are given on the unperturbed version of ScanObjectNN}
\label{table:expAblation}

\begin{center}
\begin{tabular}{|c| c| c| c| c| }
  \hline
  \textbf{Layer} & \textbf{Pooling} & \textbf{With backg.} & \textbf{Acc.} & \textbf{Cls Acc.} \\
  \hline
  GAT & MaxPool & N & 38.1 & 38.3 \\
  \hline
  KPConv & MaxPool & N & 31.4 & 33.4 \\
  \hline
  SKPConv & MaxPool & N & 52.1 & 48.3 \\
  \hline
  SKPConv & VoteMaxPool & N & 52.7 & 48.7 \\
  \hline
  SKPConv & MaxPool & Y & 44.0 & 41.3 \\
  \hline
  SKPConv & VoteMaxPool & Y & 46.3 & 43.5 \\
  \hline
\end{tabular}
\end{center}
\end{table}

%%%%%%%%%%%%%%%%%%%%%%%%%%%%%%%%%%%%%%%%%%%%%%%%%%%%%%%%%%%%%%%%%%%%%%%%%%%%%%%%
%%%%%%%%%%%%%%%%%%%%%%%%%%%%%%%%%%%%%%%%%%%%%%%%%%%%%%%%%%%%%%%%%%%%%%%%%%%%%%%%
\section{Conclusion}
We presented a novel method that reduces the Sim2Real gap in 3D object classification. This is achieved both through a better robustness to under segmentation using a voting scheme, and through spherical kernel point convolution that take advantage or relative orientation of nodes within a graph. Our conjecture that methods tested on real scenes are negatively impacted when relying entirely on a coordinate frame instead of the changes of the surface, is supported by our experimental results on the ScanObjectNN dataset~\cite{scanobjectnn-iccv19}.

The next step for this work is to investigate the applicability of similar concepts to Sim2Real object detection. Indeed, a system able to learn from a set of CAD models would provide an extremely data-efficient method to detect objects in real-world scenes.

\bibliographystyle{IEEEtranS}
\bibliography{bib}

\begin{thebibliography}{10}
\providecommand{\url}[1]{#1}
\csname url@rmstyle\endcsname
\providecommand{\newblock}{\relax}
\providecommand{\bibinfo}[2]{#2}
\providecommand\BIBentrySTDinterwordspacing{\spaceskip=0pt\relax}
\providecommand\BIBentryALTinterwordstretchfactor{4}
\providecommand\BIBentryALTinterwordspacing{\spaceskip=\fontdimen2\font plus
\BIBentryALTinterwordstretchfactor\fontdimen3\font minus
  \fontdimen4\font\relax}
\providecommand\BIBforeignlanguage[2]{{%
\expandafter\ifx\csname l@#1\endcsname\relax
\typeout{** WARNING: IEEEtran.bst: No hyphenation pattern has been}%
\typeout{** loaded for the language `#1'. Using the pattern for}%
\typeout{** the default language instead.}%
\else
\language=\csname l@#1\endcsname
\fi
#2}}

\bibitem{3dmFV-8394990}
Y.~{Ben-Shabat}, M.~{Lindenbaum}, and A.~{Fischer}, ``3dmfv: Three-dimensional
  point cloud classification in real-time using convolutional neural
  networks,'' \emph{IEEE Robotics and Automation Letters}, vol.~3, no.~4, pp.
  3145--3152, 2018.

\bibitem{bobkov2018noise}
D.~Bobkov, S.~Chen, R.~Jian, M.~Z. Iqbal, and E.~Steinbach, ``Noise-resistant
  deep learning for object classification in three-dimensional point clouds
  using a point pair descriptor,'' \emph{IEEE Robotics and Automation Letters},
  vol.~3, no.~2, pp. 865--872, 2018.

\bibitem{Gong_2019_ICCV}
S.~Gong, L.~Chen, M.~Bronstein, and S.~Zafeiriou, ``Spiralnet++: A fast and
  highly efficient mesh convolution operator,'' in \emph{Proceedings of the
  IEEE/CVF International Conference on Computer Vision (ICCV) Workshops}, Oct
  2019.

\bibitem{batchnorm_ioffe_2015}
S.~Ioffe and C.~Szegedy, ``Batch normalization: {A}ccelerating deep network
  training by reducing internal covariate shift,'' in \emph{Proceedings of
  International Conference on Machine Learning}, 2015, pp. 448--456.

\bibitem{kipf2017semi}
T.~N. Kipf and M.~Welling, ``Semi-supervised classification with graph
  convolutional networks,'' in \emph{Proceedings of International Conference on
  Learning Representations}, 2017.

\bibitem{pointcnnli2018}
Y.~Li, R.~Bu, M.~Sun, W.~Wu, X.~Di, and B.~Chen, ``Pointcnn: Convolution on
  $\chi$-transformed points,'' in \emph{Proceedings of the 32nd International
  Conference on Neural Information Processing Systems}, 2018, pp. 828--838.

\bibitem{voxnetmaturana2015}
D.~Maturana and S.~Scherer, ``Voxnet: A 3d convolutional neural network for
  real-time object recognition,'' in \emph{2015 IEEE/RSJ International
  Conference on Intelligent Robots and Systems (IROS)}.\hskip 1em plus 0.5em
  minus 0.4em\relax IEEE, 2015, pp. 922--928.

\bibitem{votenetqi2019deep}
C.~R. Qi, O.~Litany, K.~He, and L.~J. Guibas, ``Deep hough voting for 3d object
  detection in point clouds,'' in \emph{Proceedings of the IEEE/CVF
  International Conference on Computer Vision}, 2019, pp. 9277--9286.

\bibitem{qi_pointnet:_2017}
C.~R. Qi, H.~Su, K.~Mo, and L.~J. Guibas, ``Pointnet: Deep learning on point
  sets for 3d classification and segmentation,'' in \emph{Proceedings of the
  IEEE Conference on Computer Vision and Pattern Recognition}, 2017, pp.
  652--660.

\bibitem{pointnet++qi2017}
C.~R. Qi, L.~Yi, H.~Su, and L.~J. Guibas, ``Pointnet++ deep hierarchical
  feature learning on point sets in a metric space,'' in \emph{Proceedings of
  the 31st International Conference on Neural Information Processing Systems},
  2017, pp. 5105--5114.

\bibitem{iclr_2017_dl_sets_pc}
S.~Ravanbakhsh, H.~Su, J.~Schneider, and B.~Poczos, ``Deep learning with sets
  and point clouds,'' in \emph{Proceedings of International Conference on
  Learning Representations}, 2017.

\bibitem{rotkpconv9320370}
H.~{Thomas}, ``Rotation-invariant point convolution with multiple equivariant
  alignments.'' in \emph{2020 International Conference on 3D Vision (3DV)},
  2020, pp. 504--513.

\bibitem{KPConvthomas2019}
H.~Thomas, C.~R. Qi, J.-E. Deschaud, B.~Marcotegui, F.~Goulette, and L.~J.
  Guibas, ``Kpconv: Flexible and deformable convolution for point clouds,''
  \emph{Proceedings of the IEEE International Conference on Computer Vision},
  2019.

\bibitem{scanobjectnn-iccv19}
M.~A. Uy, Q.-H. Pham, B.-S. Hua, D.~T. Nguyen, and S.-K. Yeung, ``Revisiting
  point cloud classification: A new benchmark dataset and classification model
  on real-world data,'' in \emph{International Conference on Computer Vision
  (ICCV)}, 2019.

\bibitem{gatvelickovic2018graph}
P.~Veli{\v{c}}kovi{\'{c}}, G.~Cucurull, A.~Casanova, A.~Romero, P.~Li{\`{o}},
  and Y.~Bengio, ``Graph attention networks,'' in \emph{International
  Conference on Learning Representations}, 2018, (accepted as poster).

\bibitem{verma_feastnet:_2018}
N.~Verma, E.~Boyer, and J.~Verbeek, ``Feastnet: Feature-steered graph
  convolutions for 3d shape analysis,'' in \emph{Proceedings of the IEEE
  Conference on Computer Vision and Pattern Recognition}, 2018, pp. 2598--2606.

\bibitem{dgcnnwang2019}
Y.~Wang, Y.~Sun, Z.~Liu, S.~E. Sarma, M.~M. Bronstein, and J.~M. Solomon,
  ``Dynamic graph cnn for learning on point clouds,'' \emph{Acm Transactions On
  Graphics (tog)}, vol.~38, no.~5, pp. 1--12, 2019.

\bibitem{weibel2019addressing}
J.-B. Weibel, T.~Patten, and M.~Vincze, ``Addressing the sim2real gap in
  robotic 3-d object classification,'' \emph{IEEE Robotics and Automation
  Letters}, vol.~5, no.~2, pp. 407--413, 2019.

\bibitem{wohlkinger_ensemble_2011}
W.~Wohlkinger and M.~Vincze, ``Ensemble of shape functions for {3D} object
  classification,'' in \emph{Proceedings of IEEE International Conference on
  Robotics and Biomimetics}, 2011, pp. 2987--2992.

\bibitem{modelnet}
Z.~Wu, S.~Song, A.~Khosla, F.~Yu, L.~Zhang, X.~Tang, and J.~Xiao, ``{3D
  ShapeNets: A} deep representation for volumetric shapes,'' in
  \emph{Proceedings of IEEE Conference on Computer Vision and Pattern
  Recognition}, 2015, pp. 1912--1920.

\bibitem{spidercnnxu2018}
Y.~Xu, T.~Fan, M.~Xu, L.~Zeng, and Y.~Qiao, ``Spidercnn: Deep learning on point
  sets with parameterized convolutional filters,'' in \emph{Proceedings of the
  European Conference on Computer Vision (ECCV)}, 2018, pp. 87--102.

\end{thebibliography}

\end{document}